\documentclass[journal]{IEEEtran}
\usepackage{color}

\ifCLASSINFOpdf
  \usepackage[pdftex]{graphicx}
  \graphicspath{{../pdf/}{../jpeg/}}
  \DeclareGraphicsExtensions{.pdf,.jpeg,.png}
\fi
\usepackage{float}
\usepackage{makecell}
\usepackage{longtable}
\usepackage{stfloats}
\usepackage{multirow}
\usepackage{amsmath}
\usepackage{graphicx}
\usepackage{bm}
\usepackage{subfigure}
\usepackage[numbers,sort&compress]{natbib}

\usepackage{balance}

\begin{document}

\title{Deep Instance Segmentation with Automotive Radar Detection Points}

\author{\IEEEauthorblockA{Jianan Liu\IEEEauthorrefmark{1},
Weiyi~Xiong\IEEEauthorrefmark{1},
Liping~Bai,
Yuxuan~Xia,
Tao~Huang,~\IEEEmembership{Senior Member,~IEEE,} \\
Wanli~Ouyang,~\IEEEmembership{Senior Member,~IEEE,} and
Bing~Zhu\IEEEauthorrefmark{2},~\IEEEmembership{Member,~IEEE}
}
\vspace{-5 mm}

\thanks{W.~Xiong, L.~Bai and B.~Zhu were supported by National Natural Science Foundation of China under grant 62073015. 
T.~Huang was supported by the Australian Research Council Grant DP220101634. 
W.~Ouyang was supported by the Australian Research Council Grant DP200103223, Australian Medical Research Future Fund MRFAI000085, and CRC-P Smart Material Recovery Facility (SMRF) – Curby Soft Plastics.}

\thanks{J.~Liu is with Vitalent Consulting, Gothenburg 41761, Sweden, and Silo AI, Stockholm, Sweden. Email: jianan.liu@vitalent.se, jianan.liu@silo.ai.}
\thanks{W.~Xiong, L.~Bai and B.~Zhu are with School of Automation Science and Electrical Engineering, Beihang University, Beijing 100191, P.R.~China. Email: weiyixiong@buaa.edu.cn (W. Xiong);
bai\_liping@buaa.edu.cn (L. Bai);
zhubing@buaa.edu.cn (B. Zhu).}
\thanks{Y.~Xia is with Department of Electrical Engineering, Chalmers University of Technology, Gothenburg 41296, Sweden. Email: yuxuan.xia@chalmers.se.}
\thanks{T. Huang is with ICC Lab, College of Science and Engineering, James Cook University, Smithfield QLD 4878, Australia. Email: tao.huang1@jcu.edu.au.}
\thanks{W. Ouyang is with SIGMA Lab, School of Electrical and Information Engineering, The University of Sydney, Sydney NSW 2006, Australia. Email: wanli.ouyang@sydney.edu.au.}
\thanks{\IEEEauthorrefmark{1}Both authors contribute equally to the work and are co-first authors.}
\thanks{\IEEEauthorrefmark{2}Corresponding author.}
\thanks{This paper has been accepted by IEEE Transactions on Intelligent Vehicles. Digital Object Identifier 10.1109/TIV.2022.3168899}
}

\maketitle

\begin{abstract}
Automotive radar provides reliable environmental perception in all-weather conditions with affordable cost,
but it hardly supplies semantic and geometry information due to the sparsity of radar detection points.
With the development of automotive radar technologies in recent years, instance segmentation becomes possible by using automotive radar. Its data contain contexts such as radar cross section and micro-Doppler effects, and sometimes can provide detection when the field of view is obscured. 
The outcome from instance segmentation could be potentially used as the input of trackers for tracking targets. The existing methods often utilize a clustering-based classification framework, which fits the need of real-time processing but has limited performance due to minimum information provided by sparse radar detection points. In this paper, we propose an efficient method based on clustering of estimated semantic information to achieve instance segmentation for the sparse radar detection points. In addition, we show that the performance of the proposed approach can be further enhanced by incorporating the visual multi-layer perceptron. The effectiveness of the proposed method is verified by experimental results on the popular RadarScenes dataset, achieving 89.53\% mean coverage and 86.97\% mean average
precision with the IoU threshold of 0.5, which is superior to other approaches in the literature. 
More significantly, the consumed memory is around 1MB, and the inference time is less than 40ms, indicating that our proposed algorithm is storage and time efficient. These two criteria ensure the practicality of the proposed method in real-world systems.
\end{abstract}


\IEEEpeerreviewmaketitle

\section{Introduction}

\begin{figure}
    \centering
    \subfigure[Radar Detection Points]{
    \includegraphics[scale=0.9]{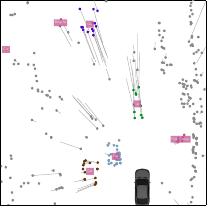}}
    \hspace{.1in}
    \subfigure[Image]{
    \includegraphics[scale=1.3]{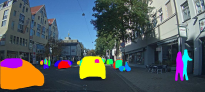}}
    \vspace{-0.1in}
    \caption{An example frame of (a) bird eye's view radar detection points and (b) its corresponding image in RadarScenes dataset \cite{radarscenes} \copyright IEEE. The measurements were collected via 77 GHz series production automotive radar sensors.}
    \label{example}
    \vspace{-5 mm}
\end{figure}

\IEEEPARstart{I}{n} the field of autonomous driving, automotive radar plays an important role in environmental perception due to its affordable cost, inherent measurement of object relative velocity, and reliability in all-weather conditions, as compared to camera and LiDAR \cite{radar_survey_1}\cite{radar_survey_2}. 
The data representation of an automotive radar is usually a set of sparse detection points generated by pre-processed raw radar signals typically in the form of a range-Doppler map or a range-azimuth heatmap.
Compared to LiDAR points, radar detection points usually provide more information, e.g., velocity (Doppler) and the radar cross section (RCS) values. However, radar detection points are much sparser and nosier than LiDAR point cloud due to their low resolution, resulting in a lack of semantic and geometric information.
Fig.~\ref{example} shows a typical scene in RadarScenes dataset \cite{radarscenes} including the collected data from both radar and camera. Note that the collected radar detection points are sparse and semantically ambiguous. Thus, it is unsuitable to directly apply methods developed for dense LiDAR point cloud to sparse radar detection points.

There are three popular methods in the literature to perform point cloud based instance segmentation \cite{PointCloudSurvey}.
The first method transforms a point cloud into a 3D grid-like representation called voxel, or projects it into a 2D grid-like representation like the bird eye's view (BEV) or a range view, and uses a convolutional neural network (CNN) to segment instance.
The drawback of this method is that it requires a large memory, high computation power, and introduces quantization error in the point-to-voxel/pixel transformation. 
The second method directly processes the points using a 1D convolutional filter-based neural network, treating spatial coordinates as part of the features.
Typical examples include the PointNets \cite{PointNet}\cite{PointNet++} and their variants \cite{SGPN}\cite{HAIS}. This method could overcome the quantization error encountered by the first method and potentially directly extracts more fruitful feature information from dense points. However, the encoders of neural networks in both first and second methods have difficulty capturing the spatial interactions of radar detection points due to their sparsity \cite{radar_detection_1}. 
The third method is to estimate which points belong to the same object using a clustering algorithm, e.g. the Density-Based Spatial Clustering of Applications with Noise (DBSCAN) \cite{DBSCAN}, and then perform the classification for each estimated cluster. Although the performance of this method is limited compared to the deep learning-based methods, it still dominates the radar-based instance segmentation in practice due to its simplicity \cite{radar_classification_1}\cite{radar_classification_2}\cite{radar_classification_3}, which minimizes the memory consumption and favors real-time data processing.

As a result, there are two aspects of challenges to perform instance segmentation on radar detection points. On the one hand, radar detection points are sparse and semantically ambiguous \cite{radar_detection_1}. Some geometric information, e.g., the shape of the object, cannot be reflected by the distribution of points, resulting in a lack of local information. Thus, global feature information can help the segmentation tasks. However, existing point cloud processing networks are mainly designed with layers that extract the local features, such as convolution layers in a CNN or set abstraction levels in PointNet++ \cite{PointNet++}. Thus, the global interactions among radar detection points cannot be extracted unless the network is deep enough. Hence, how to design a deep learning algorithm to address the sparsity and ambiguity of the radar detection points is a challenge. On the other hand, automotive driving needs an algorithm that can process the radar data in real-time, and the MCU in a radar processing system restricts the storage space of the model and available computation power. Therefore, the designed neural network must be lightweight in both size and required computation resources, which contradicts the requirement of high performance. Thus, developing a lightweight algorithm that can be used in practical scenarios while maintaining good performance is another challenge.

In this paper, we propose a new strategy and its enhancement version for instance segmentation on automotive radar detection points to combat issues in the aforementioned methods. Our goal is to develop a new and practical deep-learning architecture that can handle the sparse radar detection points for segmentation process with small memory requirement and fast run time. The main contributions of this paper are summarized as follows:
\begin{itemize}
\item We design a novel semantic segmentation-based clustering method for the instance segmentation task on sparse detection points obtained from automotive radar.
The model is designed base on the semantic segmentation version of PointNet++ \cite{PointNet++}, with a newly introduced head that estimates the point-wise center shift vector (CSV) which represents the offset in latent space from every detection point to the geometric center of its corresponding instance. 
By shifting each point toward the center of its instance with predicted CSV during clustering, the points belonging to the same instance become closer, which increase the clustering accuracy.

\item We propose to use cosine similarity (CS) loss and normalized inner product (NIP) loss in the training process of the semantic segmentation phase for sparse radar detection points, to improve the performance of CSV guided clustering.
These loss functions are designed to minimize the distance between the predicted and ground-truth CSV in latent space, resulting in more accurate prediction of CSVs.

\item We investigate how various visual multi-layer perceptrons (MLPs) \cite{external_attention}\cite{MLP-Mixer}\cite{gMLP} can be incorporated to the proposed method, and propose to employ gMLP to further improve the performance of our model. A thorough search of the literature yielded no existing study on adopting visual MLPs in radar sensing to overcome the limitation of local feature extraction due to the sparsity of radar detection points.
In addition, we propose tailored lightweight methods to achieve a balance among the computation speed, memory consumption and accuracy. 

\item We experiment with the proposed method on the recent RadarScenes dataset \cite{radarscenes} and demonstrate that the performance of the proposed method outperforms the two existing methods 
by a large margin. Precisely, the mean coverage (mCov) and the mean average precision (mAP) of our selected method are 9.0\% and 9.2\% higher than the clustering-based classification method, and 8.6\% and 9.1\% higher than the end-to-end instance segmentation method, respectively. At the same time, our method and its lightweight version still maintain the memory consumption around 1MB and inference time less than 40ms, which are feasible for automotive radar micro-center unit (MCU). Such an outcome reveals the potential to apply our proposed strategy in the real-time automotive radar perception system.
\end{itemize}

The rest of this paper is organized as follows. Section \ref{sec related} discusses the related works, including point cloud processing methods, visual MLPs, and radar-based perception. The proposed radar-based instance segmentation method and its enhancements are described in Section \ref{sec methods}. Section \ref{sec results} describes the experiment process and presents the results on the RadarScenes dataset. Discussions are also made in this section. Finally, concluding remarks are drawn in section \ref{sec conclusion}.

\section{Related Works}\label{sec related}
There are three main considerations in this work: point cloud processing, visual MLPs and automotive radar-based perception.
We will discuss each of them in this section.

\subsection{Point Cloud Processing}
Points in the point cloud are sporadic and permutation invariant, making effective information extraction challenging. While the image processing techniques, such as 2D convolution, can be extrapolated into the realm of 3D point cloud data processing, the outcomes of such approaches turn out to be ineffective. 

PointNet \cite{PointNet} and the subsequent variants \cite{PointNet++} are network structures designed specifically for point cloud data, where the input data points are projected into a higher dimension space before going through a permutation invariant function, e.g., a max pooling function, for feature extraction. As opposed to PointNet that taking all the data points as the input of the first layer, PointNet++ \cite{PointNet++} seeks to imitate the convolution layer of 2D images and attempt to capture the local context. To achieve this, a set abstraction (SA) layer is used to sample, group points and capture local structure. Furthermore, the feature propagation (FP) layer is devised to propagate features from sampled points to original points and get point-wise features, for the purpose of segmentation.

PointNet and PointNet++ do not provide the function of direct instance segmentation, but some efforts have been made toward this direction. For instance, SGPN \cite{SGPN} takes PointNets as its backbone and introduce a similarity matrix for instance segmentation. HAIS \cite{HAIS} combines PointNets with clustering, adopts hierarchical aggregation to progressively generate instance proposals.

\begin{figure*}[t]
    \centering
    \includegraphics[scale=0.54]{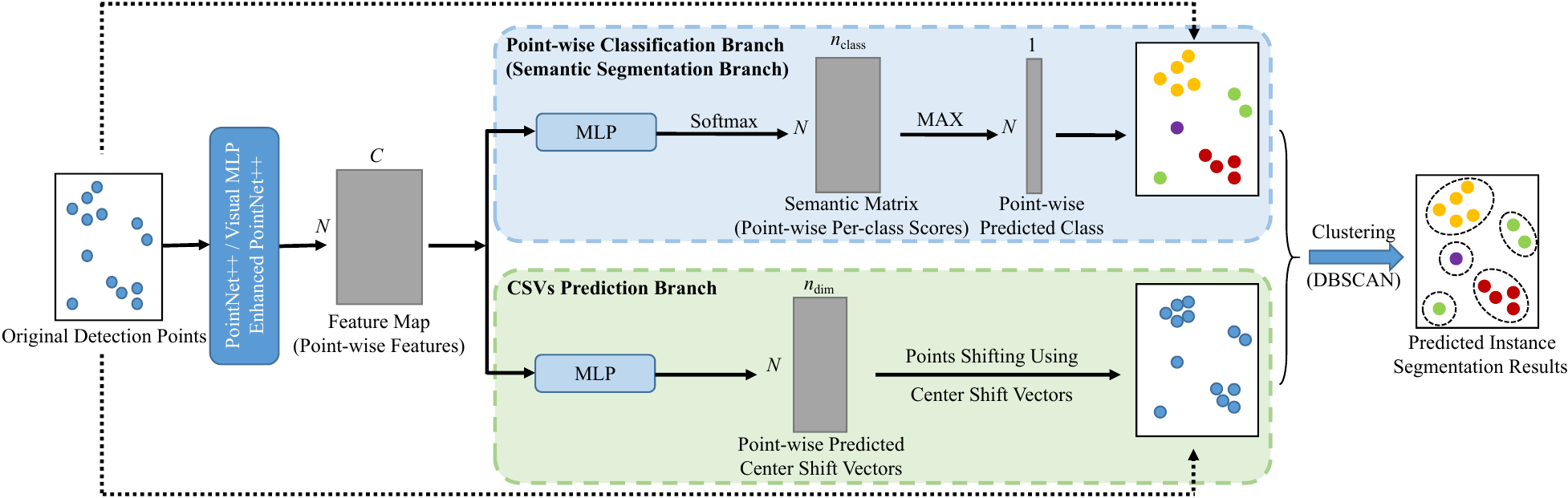}
    \caption{Illustration for the entire process of our proposed semantic segmentation-based clustering. $N$ denotes the sample size, and $C$ represents the number of output channels of the backbone; $n_{\rm{class}}$ is the number of classes and $n_{\rm{dim}}$ denotes the dimension of raw radar detection points. The input points obtain their predicted class label through the point-wise classification branch firstly. Then they are shifted according to their predicted CSVs which are estimated by the CSVs prediction branch, so that the points belonging to the same instance are more concentrated. Points with the same class label are then clustered into clusters (i.e., instances). In the instance segmentation results of the example frame, different colors denote different classes, where points in the same circle belong to the same instance.}
    \label{sem_seg_process}
    \vspace{-5 mm}
\end{figure*}

\subsection{Visual MLPs}
Attention-based transformers \cite{ViT}\cite{CSWin}\cite{SwinT} are popular approaches for computer vision tasks, but some recent works prove that comparable performance can be achieved by using MLPs only. MLP-mixer \cite{MLP-Mixer} replaces the the multi-head self attention \cite{self_attention} with a linear layer implemented on the spatial dimension. To make the MLP flexible to receive images of different sizes, researchers propose cycle-MLP \cite{CycleMLP}, where a cycle fully connected (FC) layer is used to replace the spatial MLP in MLP-mixer \cite{MLP-Mixer}. However, as the data structure of images and point cloud are different, such method cannot be applied to our work without voxelizing the point cloud. gMLP \cite{gMLP} discards the multi-head self attention in transformer and add a spatial gating unit to capture spatial interaction. Taking the inspiration from self-attention, external attention \cite{external_attention} uses two linear layers with a double normalization in between, to reduce the computational complexity.

\subsection{Automotive Radar-based Perception} 
Automotive radar-based perception, including semantic segmentation, clustering, classification, instance segmentation, object detection, and tracking, has played an essential role in the modern ADAS and autonomous driving system. With the availability of large-scale radar datasets \cite{radarscenes}, automotive radar detection points-based perception has been investigated recently \cite{radar_classification_1}\cite{radar_classification_2}\cite{radar_clustering_1}\cite{radar_semantic_segmentation}\cite{radar_detection_2}.
A two-stage clustering algorithm is designed in \cite{radar_clustering_1}. Moreover, estimated state information using an extended target tracking algorithm is employed in \cite{radar_clustering_2} as prior information to provide more stable clustering. Conventional machine learning and modern deep learning methods have also been explored for automotive radar detection points-based perception. For example, the radar detection points are used as input data for semantic segmentation task \cite{radar_semantic_segmentation}, together with occupancy grid representation of environments \cite{radar_semantic_segmentation_2}\cite{radar_semantic_segmentation_3}. Such detection points representation of radar data is also employed for classification, object detection and tracking purpose. For example, \cite{radar_classification_1}\cite{radar_classification_2}\cite{radar_classification_0} utilize random forest and LSTM to classify clustered detection points, \cite{radar_detection_1} modifies the PointNets for object detection but only one class (cars) are considered, and \cite{radar_detection_2} performs detection and tracking by adopting the combination of PointNet++ based neural network with a Kalman filter and global nearest neighbor for ID assignment over multiple frames.

\begin{figure*}
    \centering
    \includegraphics[scale=0.54]{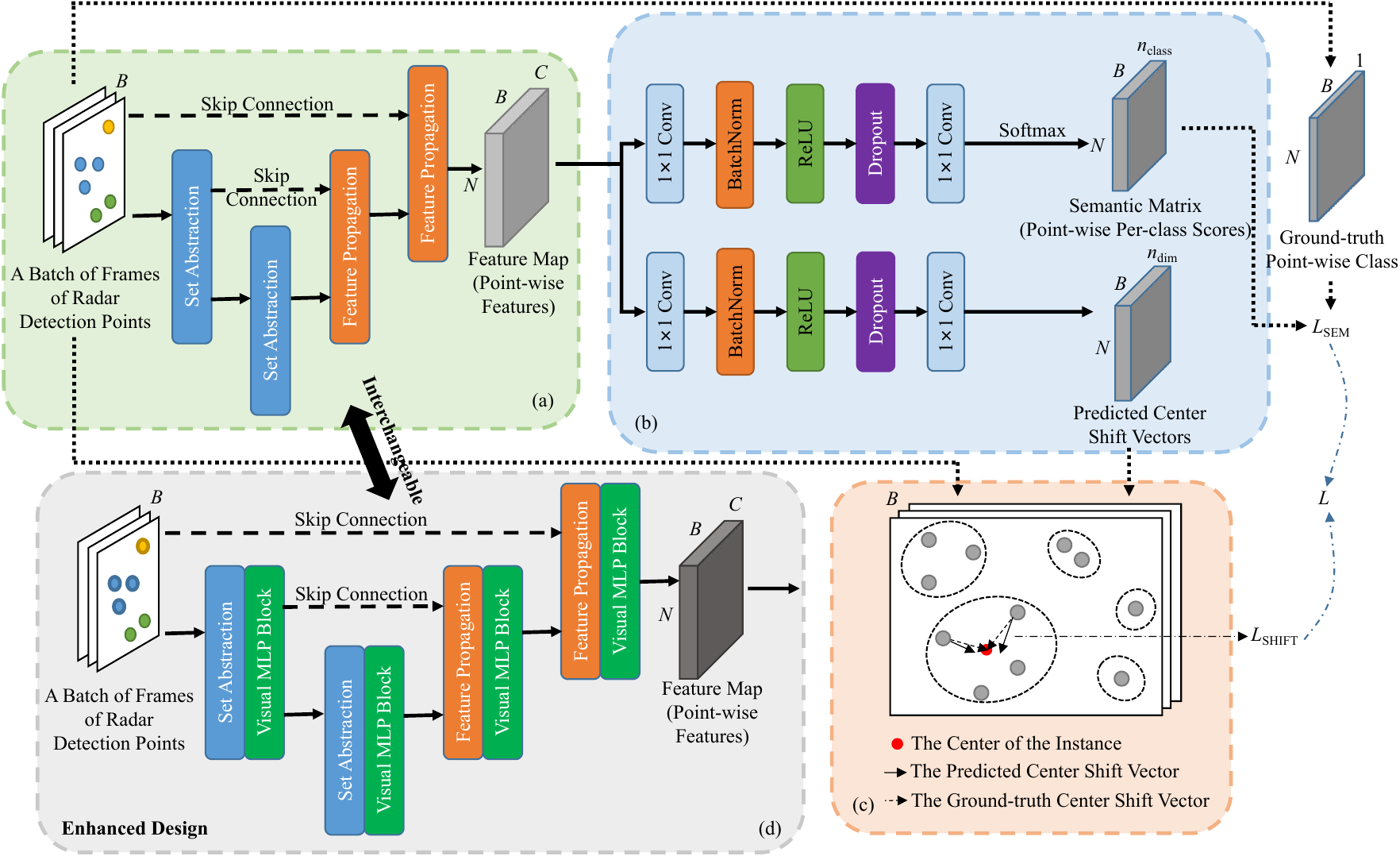}
    \vspace{-0.1in}
    \caption{The structure and the training process of our proposed semantic segmentation model. $B$ is the batch size, $N$ denotes the sample size, and $C$ represents the number of output channels of the backbone; $n_{\rm{class}}$ is the number of classes and $n_{\rm{dim}}$ denotes the dimension of raw radar detection points. (a) The PointNet++ backbone. The structure of two SA layers followed by two FP layers are applied. (b) The structure of prediction heads, which are two-layer MLPs.
    (c) The illustration of the CSVs, where points in the same circle belong to the same instance. (d) The overall structure of the visual MLP enhanced PointNet++ backbone network. The visual MLP block (e.g. gMLP \cite{gMLP}, MLP-Mixer \cite{MLP-Mixer}, external attention \cite{external_attention}, etc) is added after each set abstraction layer and feature propagation layer.}
    \label{sem_seg_model}
    \vspace{-5 mm}
\end{figure*}

\section{Proposed Methods}\label{sec methods}
In this section, we first introduce the proposed semantic segmentation-based clustering strategy. Then, we present its enhanced version with visual MLPs.

\subsection{Semantic Segmentation-based Clustering}

Dense points, such as LiDAR point cloud, on one hand has rich geometric and semantic information. On the other hand, GPUs are usually available to support processing of massive number of points in LiDAR point cloud, so that large models could be designed and high computational algorithms could be developed. In contrast, sparse radar detection points which are semantically ambiguous, cannot reflect the shape of objects. In addition, only MCU with low computational capability and small memory space could be available to process radar detection points in real time, in the typical low cost automotive radar system. As a result, the existing end-to-end instance segmentation methods designed for dense point cloud \cite{SGPN} are not suitable for sparse radar detection points. To validate this viewpoint, experiments on SGPN \cite{SGPN} are performed and compared with our proposed method. More details and discussions are in Section \ref{sec results and analysis}.

In this work, we propose an instance segmentation method for radar detection points, which is semantic segmentation (point-wise classification) based clustering.
Specifically, we first let the network model concentrate on semantic segmentation. Then we apply a clustering method for each class of detection points, since detection points assigned with different semantic information can scarcely belong to the same instance. 

Moreover, it is intuitive that clustering different classes of points with different clustering parameter settings may achieve better performance, because the attributes of clusters belong to various class types might be significantly different. 
For instance, there may be more than 10 detection points from a large vehicle, and the distances between these detection points could be larger than those from a two-wheeler, while a pedestrian may only have one detection point.
Thus, different clustering parameter settings can be used for different classes. 

Our proposed semantic segmentation-based clustering method is illustrated in Fig. \ref{sem_seg_process}, where we adopt the PointNet++ based network to perform semantic segmentation, and we choose DBSCAN as our clustering method. Concretely, the input point cloud is first processed by a PointNet++ based feature extractor to generate a feature map, and then point-wise features are sent to the point-wise classification branch to obtain the semantic information.
In consideration that simply applying DBSCAN cannot provide satisfying instance segmentation results for radar detection points due to their sparsity, we introduce a CSVs prediction branch in our network, as shown in Fig. \ref{sem_seg_process}, to estimate the offset between every point and the geometric center of corresponding ground-truth instance.
In this way, detection points can be assembled by shifting toward the center of the instance to facilitate clustering. 
To be specific, the predicted CSVs are used to push detection points towards the center of the instances in clustering process. Then DBSCAN is implemented with different clustering parameters in parallel to estimate the final instances. 

The complete training process of our proposed semantic segmentation model are shown in Fig. \ref{sem_seg_model}(a)(b)(c). Fig. \ref{sem_seg_model}(a) illustrates the PointNet++ backbone, which is composed of two SA levels and two FP levels and extracts the point-wise features; Fig. \ref{sem_seg_model}(b) shows the architecture of the two prediction heads, which are two-layer MLPs. The semantic segmentation branch predicts per-class score for every point, and for the CSVs prediction branch, the difference between each point and the center of its instance is predicted. 
An illustration of CSVs is shown in Fig. \ref{sem_seg_model}(c). 

The loss function of this new deep learning architecture can be defined as below:
\begin{equation}
    L=L_{\rm{SEM}}+\alpha L_{\rm{SHIFT}},
\end{equation}
where $L_{\rm{SEM}}$ is the cross entropy loss for semantic segmentation, $L_{\rm{SHIFT}}$ is the loss for CSVs prediction and $\alpha>0$ is a corresponding weighting factor.

Considering that the $l_2$ loss used in \cite{HAIS} only focuses on the length approximation, we propose to use the combination of the CS loss and NIP loss which minimize the difference in both angle and length between two vectors as $L_{\rm{SHIFT}}$.
More specifically, the loss for prediction of CSVs is defined by $L_{\rm{SHIFT}}=L_{\rm{CS}}+L_{\rm{NIP}}$, where $L_{\rm{CS}}$ and $L_{\rm{NIP}}$ are the CS loss and NIP loss, respectively, calculated by
\begin{align}
    L_{\rm{CS}}=&1-cosine\_similarity(\Delta \boldsymbol{x}_{\rm{pred}},\Delta \boldsymbol{x}_{\rm{gt}})\label{L_CS},\\
    L_{\rm{NIP}}=&\left|\frac{inner\_product(\Delta \boldsymbol{x}_{\rm{pred}},\Delta \boldsymbol{x}_{\rm{gt}})}{\left\|\Delta \boldsymbol{x}_{\rm{gt}}\right\|^2+\epsilon}-1\right|\label{L_NIP},
\end{align}
where $\epsilon$ is a small positive number, e.g., $10^{-5}$, to prevent singularity; $\Delta \boldsymbol{x}_{\rm{pred}}$ denotes the predicted CSV between every detection point and geometric center of corresponding instance; $\Delta \boldsymbol{x}_{\rm{gt}}$ denotes its ground truth value; {$cosine\_similarity(\cdot,\cdot)$ and $inner\_product(\cdot,\cdot)$ calculates the cosine value of the included angle and the inner product between two feature vectors, respectively. Specifically, if $\boldsymbol{x}_1$ and $\boldsymbol{x}_2$ are vectors with the same dimension, then
\[
\begin{aligned}
&cosine\_similarity(\boldsymbol{x}_1,\boldsymbol{x}_2)=\frac{\boldsymbol{x}_1^{\rm{T}}\boldsymbol{x}_2}{\left\|\boldsymbol{x}_1\right\|\left\|\boldsymbol{x}_2\right\|},\\
&inner\_product(\boldsymbol{x}_1,\boldsymbol{x}_2)=\boldsymbol{x}_1^{\rm{T}}\boldsymbol{x}_2.
\end{aligned}
\]}

As shown in \eqref{L_CS} and \eqref{L_NIP}, the included angle approaches zero due to the CS loss; the NIP loss is designed for length approximation. The above definition of $L_{\rm{SHIFT}}$ fully explores the offset shifting generated from feature vectors in the latent space by leveraging the multi-dimension physical feature information from the radar. Compared to $l_2$ loss, the proposed two loss functions for CSVs lead to a significant improvement in instance segmentation of radar detection points, which can be seen in Section \ref{sec results}.

\subsection{Enhancement with Visual MLPs} \label{sec MLP}

Due to the sparsity of radar detection points, some points may be far away from others, and SA levels of PointNet++ which capture the local information are not able to catch the interaction between them. As a result, the PointNet++ model may not be good at extracting global features for sparse radar detection points. To solve this problem, FC layers or visual transformers can be applied to our model. However, the global-extracting performance of FC layers or traditional MLPs are limited, and the model size of the visual transformers are too large, which is not suitable for radar processing systems that require real-time inference. As a promising alternative for visual transformers, visual MLPs integrate the advantages of both traditional MLPs and visual transformers that they have strong abilities and relatively small model sizes. 

\begin{figure}[t]
    \centering
    \includegraphics[scale=0.54]{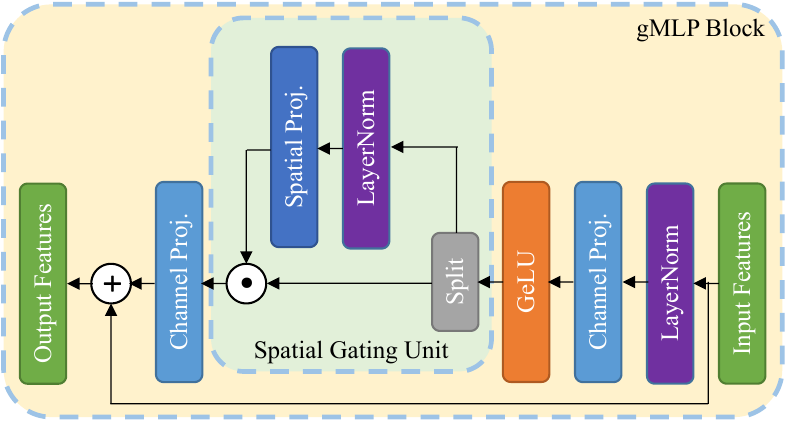}
    \vspace{-0.1in}
    \caption{The structure of the gMLP \cite{gMLP} block. The ``$\odot$'' denotes element-wise multiplication. }
    \label{gmlp_structure}
    \vspace{-5 mm}
\end{figure}

The structure of the enhanced network is shown in Fig. \ref{sem_seg_model}(d). In this paper, visual MLPs are integrated into the proposed algorithm after each SA and FP layer in PointNet++ to achieve global interaction. The extracted feature vectors are fed into a visual MLP block for down-sized feature refinement and propagated into the next layer in the encoder of the network. In the decoder, the up-sampled feature vectors are strengthened by a visual MLP block to achieve better representations in the latent space.

The visual MLP block in Fig. \ref{sem_seg_model}(d) could be any visual MLP, e.g., MLP-Mixer \cite{MLP-Mixer}, external attention \cite{external_attention}, or gMLP \cite{gMLP}. In this research, the gMLP (MLP with gating units) is adopted, whose structure is illustrated in Fig.~\ref{gmlp_structure}.

The core of gMLP is the spatial gating unit (SGU) where each feature is split into two parts along the channel dimension, one projected by a spatial projection layer and multiplied by the other. The spacial projection makes the extracted features share information with each other globally, and the element-wise multiplication retains the local features extracted by the previous modules, so the SGU can tune the extracted features according to both the global and local information. Moreover, a channel projection layer is used in the beginning and the end of the block to mix different components of each feature along the channel dimension, which makes the learned features more flexible. In general, gMLP is superior than the MLP-Mixer whose block is mainly composed of a spacial-mixing MLP and a channel-mixing MLP, as gMLP can be seen as an improved version of the MLP-Mixer by combining SGU with the latter. By incorporating gMLP into our proposed network, a beneficial mixture of local and global features can be retrieved adaptively from all points at a single frame, allowing for correct semantic guidance while taking into account differences in sparsity of radar detection points from one frame to another.

The performance of the network with gMLP as well as with other visual MLPs are compared and analyzed in Section \ref{sec results}.

\section{Experiments and Results}\label{sec results}

\subsection{Dataset}
\begin{figure}
    \centering
    \includegraphics[scale=0.25]{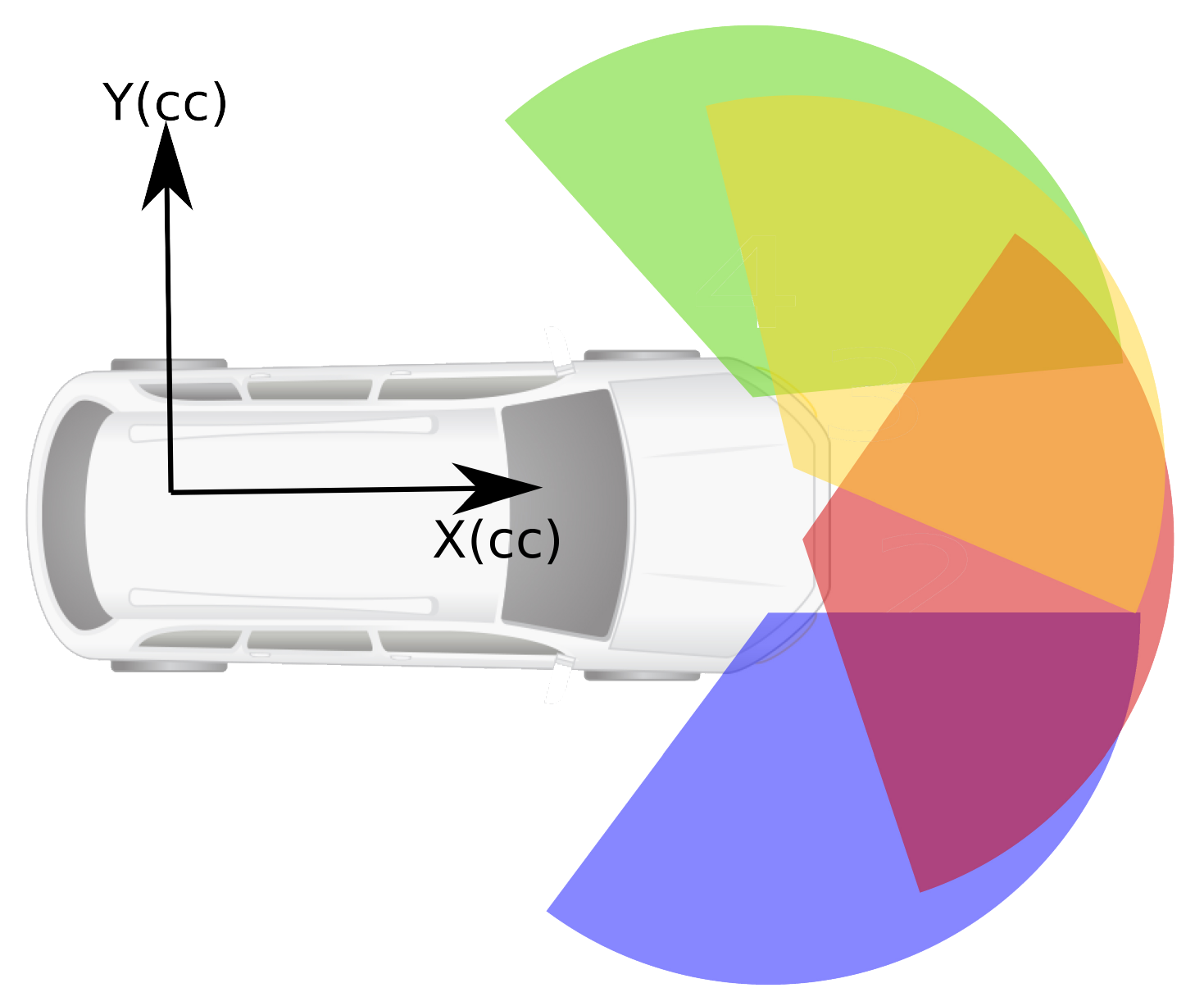}
    \vspace{-0.1in}
    \caption{The illustration of automotive radar settings in radarscenes dataset \cite{radarscenes}. The figure is reproduced from \cite{radarscenes}. The four radars are mounted at the front end of the vehicle at $85^\circ$, $25^\circ$, $-25^\circ$, and $-85^\circ$ with respect to the driver, respectively. Each radar covers a $\pm 60^\circ$ field of view.}
    \label{radar_in_radarscenes_dataset}
    \vspace{-5 mm}
\end{figure}

There are many publicly available radar datasets, such as CARRADA \cite{CARRADA}, CRUW \cite{CRUW} and Nuscenes \cite{Nuscenes}. However, to the authors' best knowledge, RadarScenes \cite{radarscenes} is the only large dataset that provides point-wise category and instance annotation. As a result, RadarScenes \cite{radarscenes} dataset is selected to validate the proposed methods. The dataset contains data from four front-mounted near-range automotive radars, one camera, and one odometer. The four radars are $77$ GHz near field automotive radar with a detection range of up to 100 meters, mounted at the front end of the vehicle at $85^\circ$, $25^\circ$, $-25^\circ$, and $-85^\circ$ with respect to the driver, respectively.
Each radar covers a $\pm 60^\circ$ field of view. 
The illustration of mounting positions of four radars and the corresponding field of view (FOV) can be seen in Fig. \ref{radar_in_radarscenes_dataset}.
The data stream is timestamped so that the ego-coordinate of any vehicle can be used as the anchor coordinate system, and information from all four radars are synchronized in one frame. The average frame rate is $17$Hz. 

The radar provides data of position, velocity, time, and ID, in both Cartesian and polar coordinates. Among these data, the followings are useful: $RCS$ in dBsm; $vr\_compensated$ in meter/second (m/s), which is the radial velocity for this detection but compensated for the ego-motion; $x\_cc$ / $y\_cc$ in meters, which is the position of the detection horizontal / orthogonal to the car in the car coordinate system (the origin is at the center of the rear-axle).

In general, RadarScenes contains real-world radar detection points from different driving environments, which are manually annotated with a class, an instance ID and other information. Specifically, there are more than 4 hours' recording in this dataset, and it is composed of 158 sequences. In order to reduce the difference among training set, validation set and test set, we randomly shuffle the frames and split them by the proportion of $8:1:1$. As the original data are sufficient to train a supervised model, no data augmentation is performed in our experiments.

\subsection{Implementation and Parameter Setting}
Parameters of all experiments are set to the same values: the batch size is $512$, the initial learning rate is $10^{-3}$, and the optimizer is Adam. The learning rate restarts every $20$ epochs with the scheduler of Cosine Annealing Warm Restarts.

Although there are 12 classes of objects in RadarScenes, we choose the settings of 5 classes, including car, pedestrian, group of pedestrians, large vehicle and two-wheeler. In this setting, all static points are not used in our experiment since we focus on detection of dynamic objects. In addition, due to lack of data in some classes, some classes of dynamic objects are merged and some are discarded. Specifically, the classes of animals and other objects are discarded; large vehicles, trucks, buses and trains are merged into the class of large vehicles, bicycles and motorized two-wheelers are merged into the class of two-wheelers, while the class of cars, pedestrians and groups of pedestrians remains unchanged.

To solve the problem that every frame has different number of points (for convenience, the number of points in $i$-th frame is denoted by $N_i$), points are sampled randomly in each frame. Some statistics are obtained such as $max(N_i)$ and $mean(N_i)$ to determine the sample size. 
In training, the sample size should be larger than most $N_i$s because $N_i$ is usually small, but it is unnecessary to be larger than $max(N_i)$ as it would increase the computational cost. However, while inferring, the sample size must be larger than $max(N_i)$, otherwise some detection points will be missing. 
In practice, the number of non-static points in a frame varies from 1 to $173$, and less than $0.4\%$ of frames have more than 100 points, so the sample size is set to $100$ in training and $200$ in testing, except the gMLP-based network, whose parameters contains the sample size, and $200$ sample size is set in both training and testing. Note that the result is not sensitive to the sample size on the condition that the sample size is larger than the number of points in the majority of frames. This is because, under this circumstance, most of the information contained in the frames can be utilized. Furthermore, the points not sampled in some epoch may be sampled in another epoch so that their information will not be lost totally. Thus, changing the sample size will not affect the performance much.

By sampling (or repeating), there exist $100$ non-static points in each frame. We approximate the FOV to a $100\mathrm{m} \times 100\mathrm{m}$ field, taking into account the configuration and technical specification of the four near field radars. Based on these numerical structures, we require that the PointNet++ segmentation network \cite{PointNet++} has two SA levels and two corresponding FP levels, as illustrated in Fig.~\ref{sem_seg_model}(a). The number of sampled points in the first SA level is set to $64$ with radius $8$m. These parameters are designed such that all the sampling cycles would cover the entire FOV with appropriate overlapping:
\[
    n\pi r^2>S_{\rm{FOV}},
\]
where $n$ is the number of sampled points; $r$ denotes the radius; and $S_{\rm{FOV}}$ is the area of FOV.

The density of data points at each frame is $300/64=3.125$; therefore, there are on average $3.1$ data points in each sampling cycle. The sampling number is set to $8$ which is larger since the maximum pooling operation of the PointNet++ network is duplication insensitive and we want to guarantee no under-sampling. This design logic is extended to the second SA level, only with the input data being reduced to $64$.

In practice, the same network structure is applied to both the end-to-end instance segmentation baseline and our semantic segmentation model:
\[
\begin{aligned}
    &SA(64, 8, [8, 32, 64]),\\
    &SA(16, 16, [64, 128, 256]),\\
    &FP(64, 32),\\
    &FP(32, 32, 16),
\end{aligned}
\]
where the notations are the same as those in PointNet++ \cite{PointNet++}. Specifically, a set abstraction level is denoted by $SA(K, r, [l_1, \cdots, l_d])$, where $K$ points are sampled, and the grouping radius is $r$ for each sampled point, followed by a PointNet of $d$ 1$\times$1 convolution layers whose output channels are $l_1, \cdots, l_d$, respectively; a feature propagation level with $d$ 1$\times$1 convolution layers is represented by $FP(l_1, \cdots, l_d)$. BatchNorm and ReLU are used between two consecutive convolution layers.

For the enhanced models, the visual MLP block does not change the dimension of output feature vectors and thus it could be appended in following of each $SA$ or $FP$ directly. For the gMLP enhanced model, the first channel projection layer of each gMLP block doubles the number of channels, while the spacial projection layer and the second channel projection layer do not change the spacial or channel dimension.

\begin{figure}[t]
    \centering
    \includegraphics[scale=0.39]{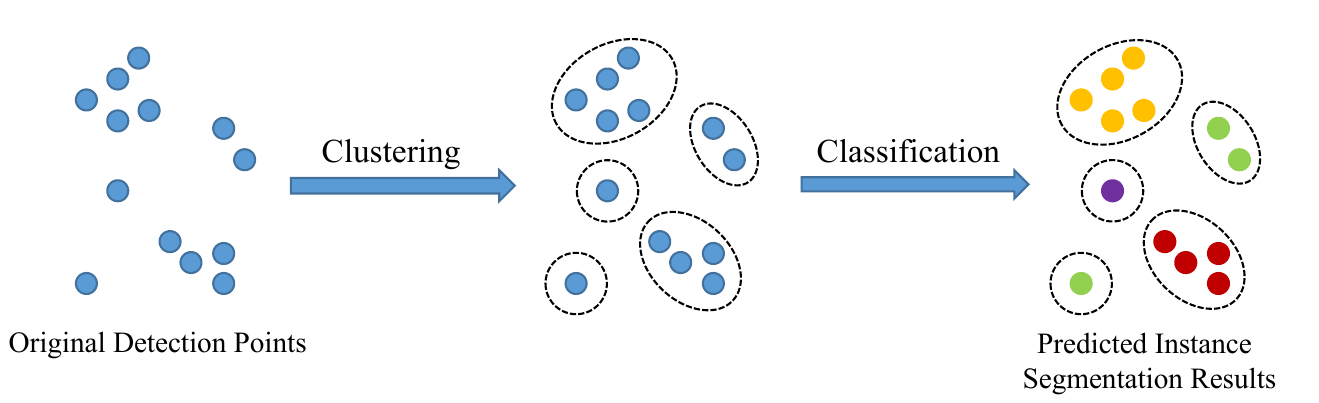}
    \vspace{-0.1in}
    \caption{Illustration for the process of clustering-based classification method. The input points are first clustered and a classifier is applied to predict a class for each cluster. Different colors denote different classes, where points in the same circle belong to the same instance.}
    \label{baseline_process}
    \vspace{-5 mm}
\end{figure}

\begin{figure*}[b]
    \centering
    \vspace{-3 mm}
    \includegraphics[scale=0.54]{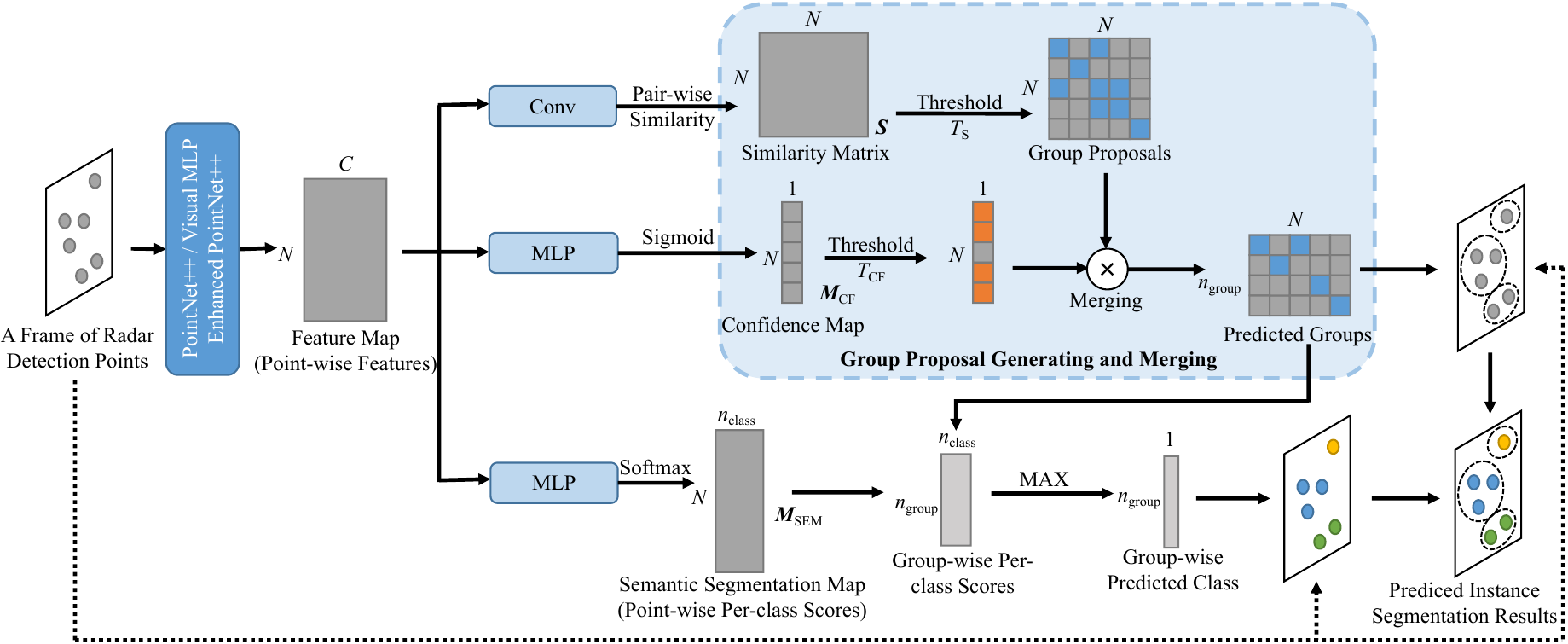}
    \vspace{-0.1in}
    \caption{Illustration for process of SGPN-based end-to-end instance segmentation. $N$ denotes the sample size, and $C$ represents the number of output channels of the backbone; $n_{\rm{class}}$ is the number of classes and $n_{\rm{group}}$ denotes the number of predicted groups in this frame. In the instance segmentation results of the example frame, different colors denote different classes, where points in the same circle belong to the same instance. For detailed information of SGPN, please refer to \cite{SGPN}.}
    \label{ins_seg_process}
\end{figure*}

The structures of different heads for our semantic segmentation model are as follows:
\[
\begin{aligned}
    &\boldsymbol{M}_{\rm{SEM}}=conv(conv(\boldsymbol{F}, 16), n_{\rm{class}}),\\
    &\boldsymbol{M}_{\rm{SHIFT}}=conv(conv(\boldsymbol{F}, 16), n_{\rm{dim}}),
\end{aligned}
\]
where $\boldsymbol{M}=conv(\boldsymbol{X}, l)$ represents a $1\times1$ convolution layer whose input is $\boldsymbol{X}$ and output is $\boldsymbol{M}$ with $l$ channels. The output of PointNet++ backbone is denoted by $\boldsymbol{F}$.
$\boldsymbol{M}_{\rm{SEM}}$ and $\boldsymbol{M}_{\rm{SHIFT}}$ are the semantic matrix and the predicted CSVs, while $n_{\rm{dim}}$ denotes the dimension of raw radar detection points. 
BatchNorm, ReLU and Dropout are used between two consecutive convolution layers.

\subsection{Existing Methods}
To show the superiority of our approach, we also experiment on two existing methods. One is the widely accepted clustering-based classification strategy for automotive radar-based instance segmentation, and the other is a deep learning-based end-to-end instance segmentation method which is originally designed for dense point cloud.

\subsubsection{Clustering-based Classification}

The commonly-used method for radar-based instance segmentation is clustering-based classification \cite{radar_classification_1}\cite{radar_classification_2}\cite{radar_classification_3}. As illustrated in Fig. \ref{baseline_process}, the input detection points are first clustered, and then each cluster is sent to a classifier, e.g., support vector machine or random forest classifier, to get the predicted class.

In our experiments, DBSCAN is chosen as the method for clustering, and random forest classifier is used to predict scores for all classes. The hand-crafted features of every cluster estimated from DBSCAN are used as the input of the random forest classifier, including the mean values and deviations of range, azimuth, Doppler and RCS.

\subsubsection{End-to-End Instance Segmentation}

In this method, the instance segmentation is implemented in a deep learning framework, and the model is built upon SGPN \cite{SGPN}, as illustrated in Fig.~\ref{ins_seg_process}. This model connects three heads to PointNet++, and predicts a similarity matrix $\boldsymbol{S}$ to estimate the possibility that any pair of the points belong to the same instance, a similarity confidence map $\boldsymbol{M}_{\rm{CF}}$ to estimate the uncertainty of similarity results, and a semantic segmentation map $\boldsymbol{M}_{\rm{SEM}}$ to provide the semantic information estimation of every point.

As shown in Fig.~\ref{ins_seg_process}, during inference, a threshold $T_{\rm{S}}$ is applied on the similarity matrix to get the group proposals, and only group proposals whose predicted confidence is greater than another threshold $T_{\rm{CF}}$ are retained. After that, group proposals with IoU larger than $T_{\rm{IoU}}$ are merged into one, and finally the predicted instances are obtained. To further determine the class of the $i$-th predicted instance $I_i$, the per-class scores of all points in $I_i$ are averaged and the class with the highest score is defined as the predicted class of $I_i$.

\newcommand{\tabincell}[2]{\begin{tabular}{@{}#1@{}}#2\end{tabular}}
\begin{table*}[]
\centering
\caption{Results of Different Strategies for Instance Segmentation on Test Data}
\label{results}
\vspace{-0.1in}
\begin{tabular}{ccccccc}
\hline
Method & \multicolumn{2}{c}{Model} & mCov(\%) & mAP$_{0.5}$(\%) & \#Params/Memory & Inference Time\\ \hline
\tabincell{c}{Clustering-based\\Classification}
& \tabincell{c}{DBSCAN + Random\\Forest Classifier}     & -             & 79.54 & 76.09 & - / - &17.1ms \\ \hline
\multirow{2}{*}{\tabincell{c}{End-to-End\\Instance Seg.}}  &\multirow{2}{*}{\tabincell{c}{SGPN}}
& $l_2$ Loss for $\boldsymbol{M}_{\rm{CF}}$                   & 77.32 & 73.21 & 75.8K/0.326MB & 63.8ms \\
& & Our BCE Loss for $\boldsymbol{M}_{\rm{CF}}$             & 79.91 & 76.15 & 75.8K/0.326MB & 63.8ms \\ \hline
\multirow{3}{*}{\tabincell{c}{Semantic Seg.\\based Clustering}} &\multirow{3}{*}{\tabincell{c}{PointNet++\\+ DBSCAN}}
& Without CSV Head                                & 82.21 & 77.96 &75.2K/0.320MB &27.4ms \\
& & With CSV Head, $l_2$ Loss    & 82.38 & 78.17 & 75.6K/0.324MB &28.7ms \\
& & With CSV Head, CS\&NIP Loss  & \textbf{82.78} & \textbf{79.38} & 75.6K/0.324MB & 28.7ms \\ \hline
\vspace{-5 mm}
\end{tabular}
\end{table*}

\begin{table*}[]
\centering
\caption{Results of Different Enhanced Models on Test Data}
\label{enhanced_results}
\vspace{-0.1in}
\begin{tabular}{ccccccc}
\hline
Method & Model & Enhancement & mCov(\%) & mAP$_{0.5}$(\%) & \#Params/Memory & Inference Time \\ \hline
\multirow{2}{*}{\tabincell{c}{End-to-End\\Instance Seg.}} & \multirow{2}{*}{\tabincell{c}{SGPN (Our BCE\\Loss for $\boldsymbol{M}_{\rm{CF}}$)}} & - & 79.91 & 76.15 &75.8K/0.326MB &63.8ms \\
 & & gMLP & 84.11 & 81.09 &339.9K/1.346MB &336.3ms \\ \hline
 \multirow{5}{*}{\tabincell{c}{Semantic Seg.\\based Clustering}} &\multirow{5}{*}{\tabincell{c}{PointNet++ (with\\CSV Head, CS\&NIP\\ Loss) + DBSCAN}} & - & 82.78 & 79.38 &75.6K/0.324MB &28.7ms \\
 & & gMLP & \textbf{88.54} & \textbf{85.24} &339.7K/1.306MB &32.5ms \\
 & & aMLP & \textbf{89.53} & \textbf{86.97} & 435.0K/1.714MB & 35.4ms \\
 & & External Attention & 85.23 & 81.41 &122.7K/0.507MB &31.2ms \\
 & & Self Attention & 85.85 & 82.32 &218.2K/0.876MB &32.0ms \\ \hline
\vspace{-5 mm}
\end{tabular}
\end{table*}

In the training process, the loss functions of $\boldsymbol{S}$ and $\boldsymbol{M}_{\rm{SEM}}$ remain as double hinge loss and cross entropy loss. However, the binary cross entropy loss (BCE), instead of the mean square error (MSE, or $l_2$), is set as the loss function of $\boldsymbol{M}_{\rm{CF}}$ to facilitate uncertainty modelling, i.e.,
\begin{equation}
\begin{aligned}
    L_{\rm{CF}}=&-\frac{1}{N}\sum_{i=1}^N[IoU(g_i,p_i)\times\log(\boldsymbol{M}_{\rm{CF},\it{i}}) \\
    &+(1-IoU(g_i,p_i))\times\log(1-\boldsymbol{M}_{\rm{CF},\it{i}})],
\end{aligned}
\end{equation}
where $N$ is the number of points in a frame; $\boldsymbol{M}_{\rm{CF},\it{i}}$ is the $i$-th row of confidence map; and $IoU(g_i,p_i)$ is the intersection over union (IoU) between the $i$-th predicted group and the corresponding ground truth group.

\subsection{Results and Analysis} \label{sec results and analysis}
Experiments on RadarScenes are performed to evaluate the effectiveness of the proposed strategies. The spatial coordinates, velocities (compensated) and RCS values are used as inputs, while mean coverage (mCov) and mean average precision (mAP) with the IoU threshold of $0.5$ (mAP$_{0.5}$) on original detection points are reported for final instance prediction.

Table \ref{results} presents results of two existing methods and our strategy without visual MLP enhancement. It should be noted that both the number of parameters and inference time for all models are given, and the inference time is defined as the average time cost when performing instance segmentation on test dataset using a CPU. The performance of the SGPN-based end-to-end instance segmentation improves $3\%$ by modifying its loss function from $l_2$ loss to BCE loss. However, such deep learning-based end-to-end instance segmentation only outperforms the clustering-based classification method with limited improvements, but its inference time increases a lot. In contrast, the semantic segmentation-based clustering strategy reaches the mCov at $82.21\%$ and mAP at $77.96\%$ without significantly increasing inference time, and further improvement can be obtained by adding a CSVs prediction branch to PointNet++ using CS and NIP loss to achieve $82.78\%$ mCov and $79.38\%$ mAP, proving that our method is suitable for instance segmentation on sparse radar detection points. We notice that the mCov does not improve as much as the mAP when CSV head with CS\&NIP loss is applied. The phenomenon can be explained as follows: We observed that the predicted CSVs improve the performance significantly only in a small proportion of situations where the plain semantic segmentation based clustering cannot work well. Therefore, after shifting points by their predicted CSVs, the Coverage of a minority of instances increases, but the increment is limited for the mCov due to the average operation. However, the mAP can increase a lot because one main influencing factor is the number of True-Positive predictions, which could also be increased by shifting points toward the center of their instances and clustering the shifted points by DBSCAN.

\begin{figure}
    \centering
    \subfigure{\includegraphics[scale=0.96]{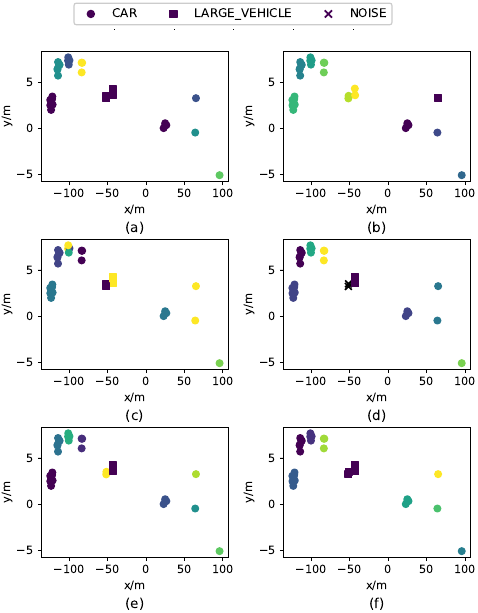}}
    \vspace{-0.1in}
    \caption{Visualization of typical instance segmentation results by using different strategies. Different shapes represent different semantics, and different colors in the same class (same shape) denotes different instances. (a) Ground Truth for instance segmentation. (b) Instance segmentation results using the strategy of clustering-based classification. (c) Instance segmentation results using the strategy of end-to-end instance segmentation. (d) Instance segmentation results using the strategy of end-to-end instance segmentation with gMLP. (e) Instance segmentation results using the strategy of semantic segmentation-based clustering. (f) Instance segmentation results using the strategy of gMLP enhanced semantic segmentation-based clustering.}
    \label{visualization}
    \vspace{-5 mm}
\end{figure}

\begin{figure*}
    \centering
    \includegraphics[scale=0.96]{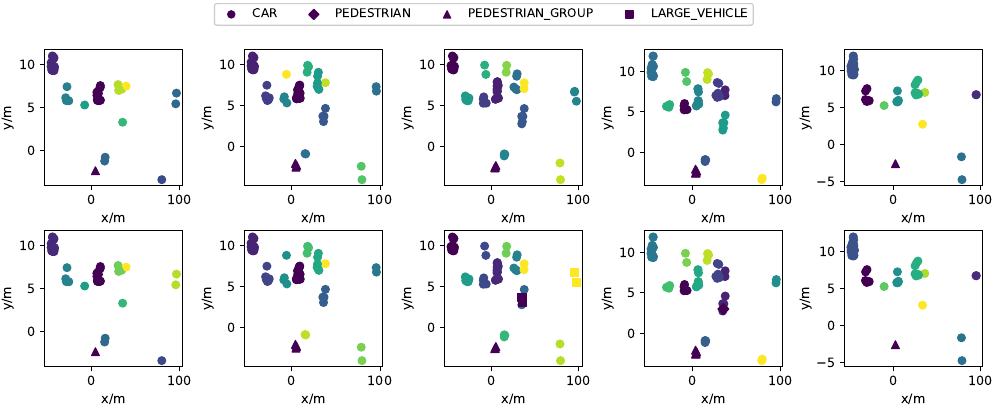}
    \vspace{-0.1in}
    \caption{Visualization of typical instance segmentation results in several consecutive frames. Strategy of gMLP enhanced semantic segmentation-based clustering is used to generate these figures. The first row is the ground truth for instance segmentation, and the second row is the predicted instance segmentation results. Note that even if there are some mistakes in the 2nd, 3rd and 4th frame, the instance segmentation results are correct in most cases.}
    \label{visualization_sequence}
    \vspace{-3 mm}
\end{figure*}

\begin{table*}
\centering
\caption{Comparison of Light-Weight Enhanced Models and Original Enhanced Models on Test Data}
\label{light_weight_enhanced_results}
\vspace{-0.1in}
\begin{tabular}{ccccccc}
\hline
Methods & Model & Enhancement \& Light-Weight Method & mCov(\%) & mAP$_{0.5}$(\%) & \#Params/Memory & Inference Time \\ \hline
 \multirow{4}{*}{\tabincell{c}{Semantic\\Seg. based\\Clustering}} & \multirow{4}{*}{\tabincell{c}{PointNet++ (with\\CSV head, CS\&NIP\\ loss) + DBSCAN}}
 & gMLP Enhancement (Group Conv) & \textbf{86.71} & \textbf{82.47} &273.0K/1.090MB &32.2ms \\
 & & gMLP (Reduced Dimension) Enhancement & \textbf{86.36} & \textbf{82.68} & 179.0K/0.731MB & 32.0ms \\
 & & External Attention Enhancement & 85.23 & 81.41 &122.7K/0.507MB &31.2ms \\
 & & Self Attention Enhancement & 85.85 & 82.32 &218.2K/0.876MB &32.0ms \\ \hline
\vspace{-5 mm}
\end{tabular}
\end{table*}

The experimental results of the enhanced models are summarized in Table \ref{enhanced_results}. It can be seen that all models in Table \ref{results} and Table \ref{enhanced_results} occupy from less than $1$MB to $2$MB storage space, making our proposed algorithm feasible for embedded radar-based perception systems. To show the effectiveness of our proposed semantic segmentation-based clustering with gMLP enhancement strategy, we use external attention and the self attention mechanism (although not an MLP) for semantic segmentation network for comparison. Comparing with its original version, the gMLP-based model performs better in mCov and mAP by approximately $6\%$, while the self attention and external attention-based model only improve about $3\%$ and $2\%$, respectively, and with similar inference time and number of parameters. 
Notably, by attaching a tiny attention module to the spatial gating unit in gMLP, such model (called aMLP) further improves performance and slightly increases the inference time and the number of parameters. 
The significant improvement after attaching gMLP/aMLP to PointNet++ can be attributed to the SGU structure of gMLP, which verifies our standpoint in Section \ref{sec MLP}. In the SGU, features can be shared globally through the spatial projection layer. As sparse radar points lack the geometric information which is part of local features, global representations are critical for classifying and associating them with the right objects. Moreover, the local and global features can be combined with flexibility due to the channel projection layers, facilitating representation learning.
For SGPN-based end-to-end instance segmentation, gMLP outperforms the baseline by approximately $5\%$, but its inference time increases dramatically, making the gMLP enhancement of such strategy infeasible in practice, which supports our idea that instance segmentation models devised for dense point cloud cannot handle the sparse points appropriately. Compare the gMLP enhanced end-to-end instance segmentation strategy with the gMLP enhanced semantic segmentation-based clustering strategy, it is witnessed that the latter outperforms the former significantly in either mCov, mAP, or inference time, indicating that only semantic segmentation-based clustering with gMLP/aMLP enhancement strategy leads to the best performance in instance segmentation while still maintains the inference time and number of parameters fairly low.

Typical examples of instance segmentation results by using different strategies is visualized in Fig. \ref{visualization}. It is clear that the clustering-based classification method might obtain the incorrect instance and semantic estimation, while the deep learning-based end-to-end instance segmentation strategy and its gMLP enhancement could correct the estimation partly, but they also generate other improper predictions. Comparatively, our semantic segmentation-based clustering method provides better estimation and its gMLP enhancement achieves $100\%$ correct prediction for this particular case. Another result of consecutive frames by using the proposed semantic segmentation-based clustering with gMLP enhancement could be seen in Fig. \ref{visualization_sequence}, where most of the instances could be segmented perfectly, even though some of them are spatially close with each other.

However, it can be also observed that two instances of cars are grouped together on the 2nd frame, and another car instance is recognized as a large vehicle on 3rd frames. Such issues could be potentially solved by incorporating the consistent information from consecutive frames. For example, a car identified at the previous frame should not be recognized as a pedestrian in the current frame.

Although semantic segmentation-based clustering with gMLP enhancement could provide acceptable storage size, it is still possible to compress it further, providing flexibility for allocating more powerful tracking algorithms in the entire radar perception system. Table \ref{light_weight_enhanced_results} shows the comparison of results of gMLP enhanced semantic segmentation model with two different compressed approaches (one uses group convolution instead of conventional convolution in PointNet++ and the other reduces the dimension of the input of spatial gating units in gMLP blocks) and other enhanced semantic segmentation models. The compressed methods are capable of reducing the number of parameters and the corresponding storage memory consumption, and it can be seen that even the compressed gMLP-based model performs better than other enhanced models with comparable or even less memory consumption.

\section{Conclusion}\label{sec conclusion}
A strategy is proposed in this paper for sparse radar detection points based instance segmentation, i.e., semantic segmentation-based clustering by PointNet++ and DBSCAN. The strategy provides better performance and faster inferring rate than the end-to-end instance segmentation model designed for dense points. Compared to the clustering-based classification method, the inference time of the latter method does not significantly increase, whereas its performance is superior.

An enhancement with gMLP/aMLP is introduced after tuning the model parameters and the loss functions. The gMLP/aMLP enhanced semantic segmentation-based clustering can provide improvements in mCov and mAP, whereas the inference time only increases slightly compared to that without gMLP.
The requirement of storage space for the enhanced model is less than $2$MB, such that the gMLP enhanced semantic segmentation-based clustering strategy is feasible for the real-time embedded radar-based ADAS/AD product. The storage space can be reduced even further, and the proposed method still maintains performance of instance segmentation by applying the light-weight approaches, indicating that more flexibility in designing radar-based perception can be achieved.

Taking detection points of several consecutive frames simultaneously would possibly enable the model to extract features with consistent information, and facilitate the learning process. This will be included in our future research.

\ifCLASSOPTIONcaptionsoff
  \newpage
\fi

\vspace{-10 mm}
\begin{IEEEbiography}[{\includegraphics[width=1in,height=1.25in,clip,keepaspectratio]{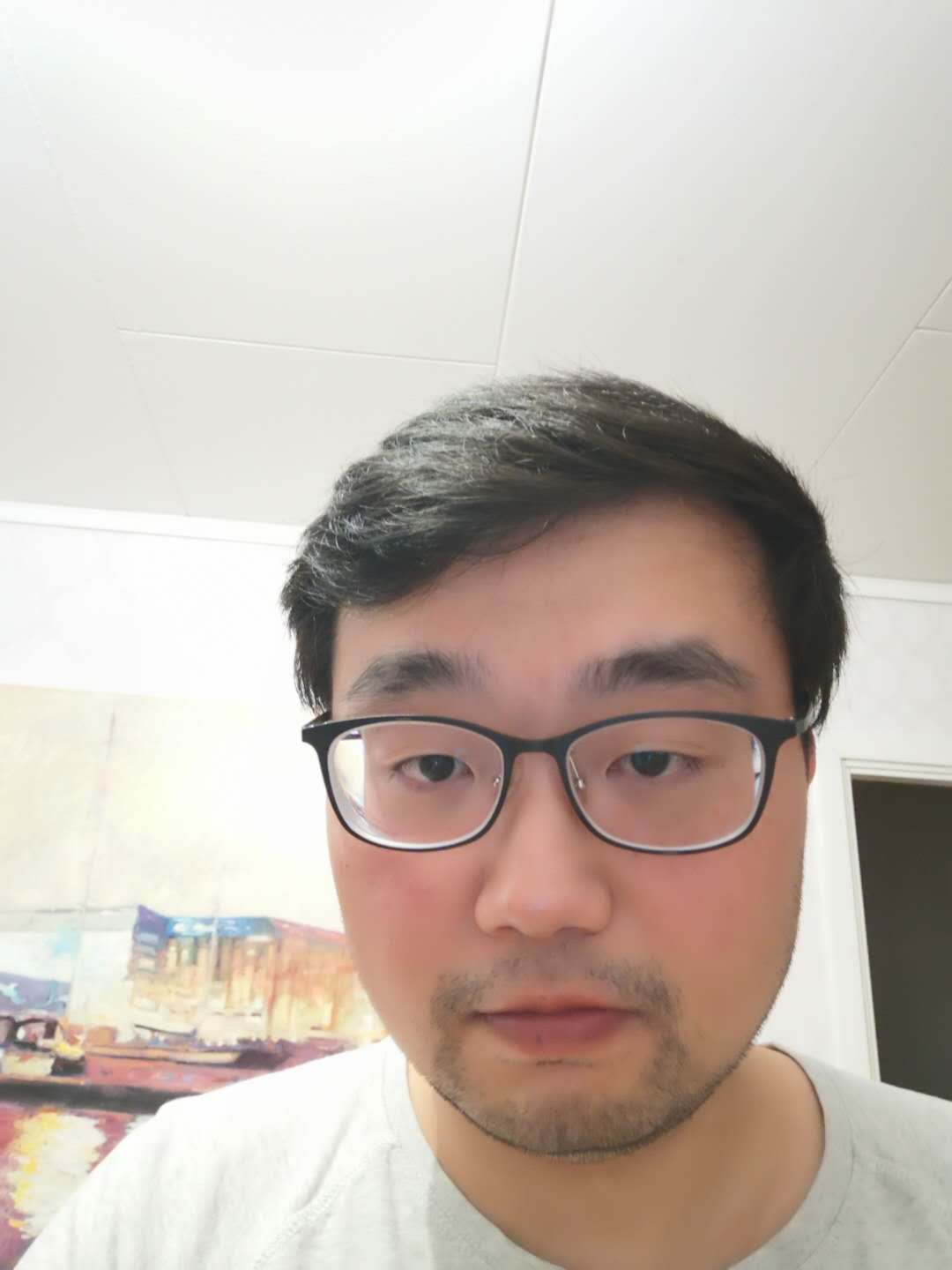}}]{Jianan Liu} received his B.Eng. degree in Electronics and Information Engineering from Huazhong University of Science and Technology, Wuhan, China, in 2007. He received his M.Eng. degree in Telecommunication Engineering from the University of Melbourne, Australia, and his M.Sc. degree in Communication Systems from Lund University, Sweden, in 2009 and 2012, respectively. 
Jianan has over ten years of experience in software and algorithm design and development. He has held senior R\&D roles in the AI consulting, automotive, and telecommunication industries. 
His research interests include applying statistical signal processing and deep learning for medical image processing, wireless communications, IoT networks, indoor sensing, and outdoor perception using a variety of sensor modalities like radar, camera, LiDAR, WiFi, etc.
\end{IEEEbiography}

\vspace{-10 mm}
\begin{IEEEbiography}[{\includegraphics[width=1in,height=1.25in,clip,keepaspectratio]{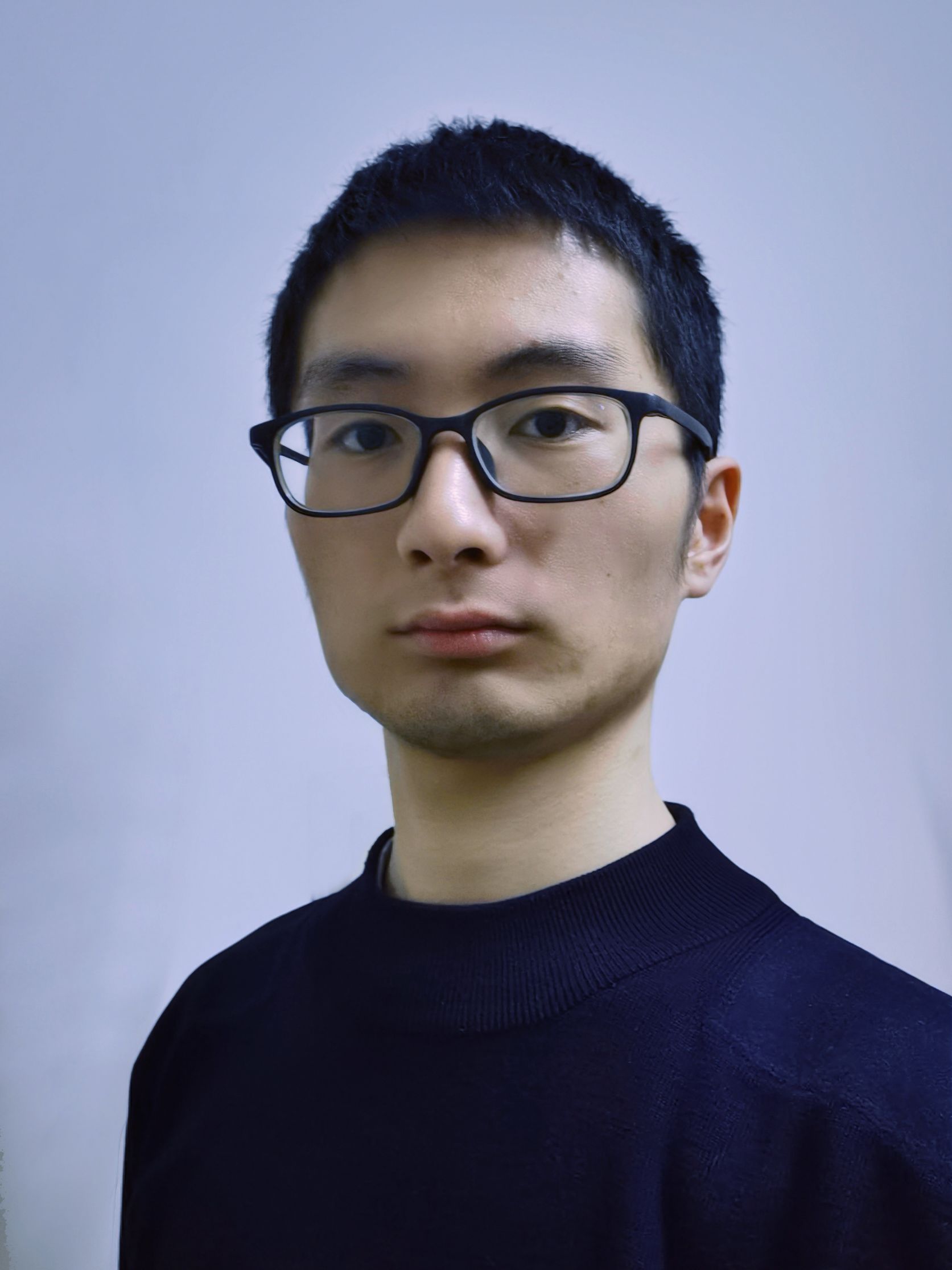}}]{Weiyi Xiong} received the B.Sc. degree in Automation from Beihang University, Beijing, China, in 2021.
He is currently working toward the M.Sc. degree in Control Science and Engineering with the School of Automation Science and Electrical Engineering, Beihang University, Beijing, China.
His research interests include deep learning, radar perception, and autonomous driving.
\end{IEEEbiography}

\vspace{-10 mm}
\begin{IEEEbiography}[{\includegraphics[width=1in,height=1.25in,clip,keepaspectratio]{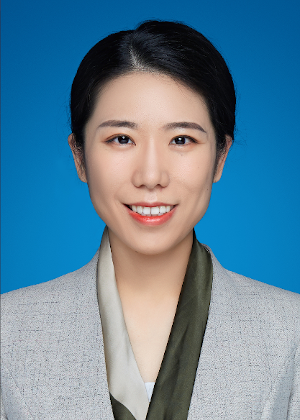}}]{Liping Bai} received her M.Eng. degree in Instrument Engineering from Nanjing University of Posts and Telecommunications, Nanjing, China, in 2021. She is currently pursuing her Ph.D. in Automation and Control Engineering at Beihang University, Beijing, China. 
Her research interest includes deep learning, automation and control theory. 
\end{IEEEbiography}

\begin{IEEEbiography}[{\includegraphics[width=1in,height=1.25in,clip,keepaspectratio]{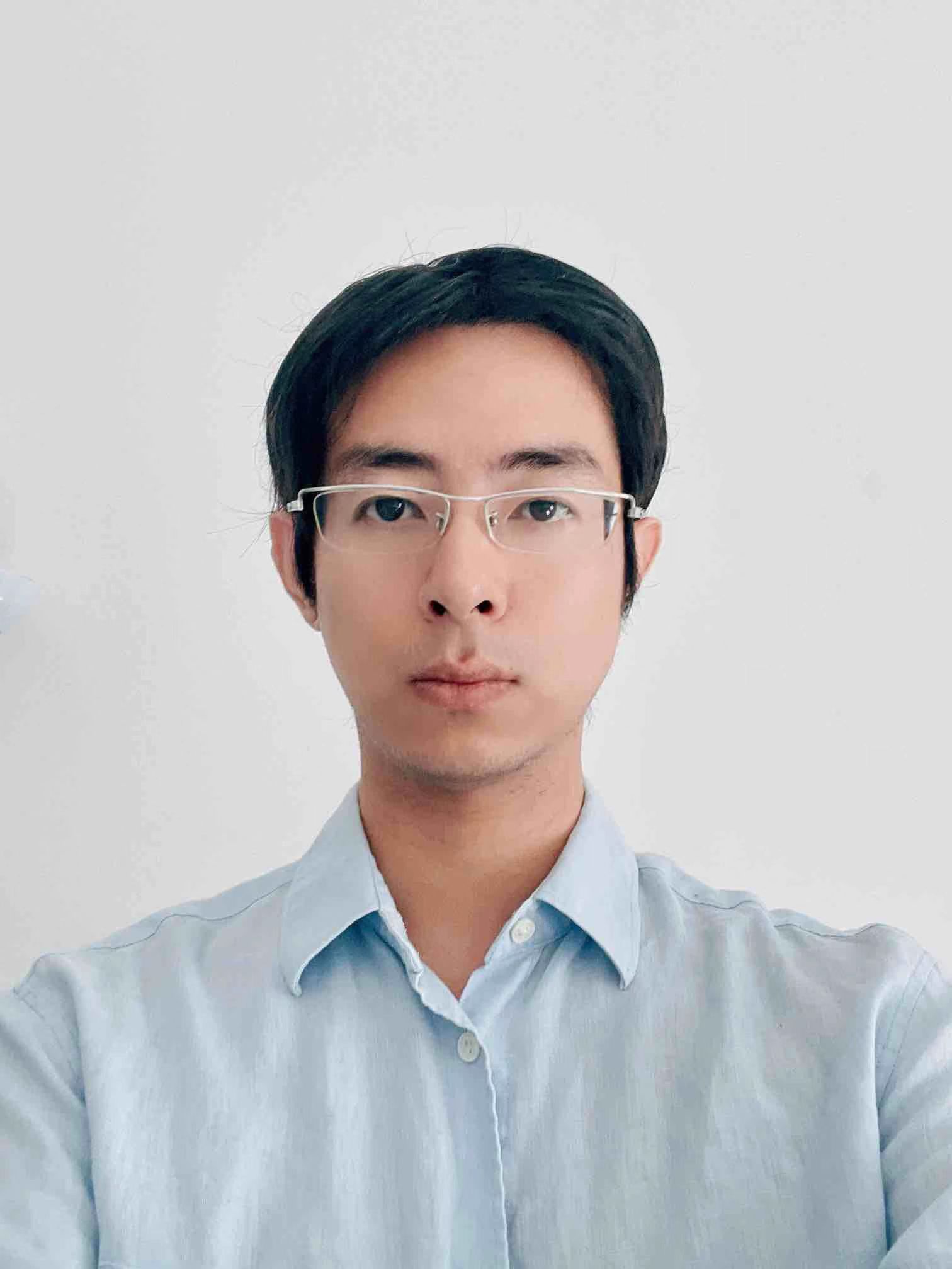}}]{Yuxuan Xia} was born in Wuhan, China, in 1993. He received the B.Sc. degree in engineering of Internet of Things from Jiangnan University, Wuxi, China, in 2015 and the M.Sc. degree in communication engineering in 2017 from the Chalmers University of Technology, Gothenburg, Sweden, where he is currently working toward the Ph.D. degree with the Department of Electrical and Engineering. His main research interests include multi-object tracking and sensor fusion, especially for extended objects. He has co-organized tutorials on multi-object tracking at Information Fusion conference.
\end{IEEEbiography}

\vspace{-10 mm}
\begin{IEEEbiography}[{\includegraphics[width=1in,height=1.25in,clip,keepaspectratio]{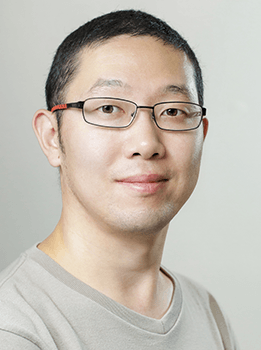}}]{Tao Huang}
(SM'20) received his Ph.D. degree in Electrical Engineering from The University of New South Wales, Sydney, Australia. 
Currently, Dr Huang is a lecturer in Electronic Systems and IoT Engineering at James Cook University, Cairns, Australia. 
He was an Endeavour Australia Cheung Kong Research Fellow, a visiting scholar at The Chinese University of Hong Kong, a research associate at the University of New South Wales, and a postdoctoral research fellow at James Cook University. 
He has co-authored a Best Paper Award from the 2011 IEEE WCNC, Cancun, Mexico.
He is currently serving as the MTT-S/Com Vice-Chair and Young Professionals Representative for the IEEE Northern Australia Section.
His research interests include deep learning, smart sensing, computer vision, pattern recognition, wireless communications, and IoT security.
\end{IEEEbiography}

\vspace{-10 mm}
\begin{IEEEbiography}[{\includegraphics[width=1in,height=1.25in,clip,keepaspectratio]{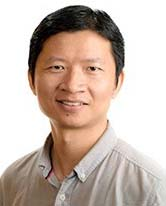}}]{Wanli Ouyang} (Senior Member, IEEE) received the Ph.D. degree from the Department of Electronic Engineering, The Chinese University of Hong Kong (CUHK). He is currently an Associate Professor with the School of Electrical and Information Engineering, The University of Sydney, Sydney, Australia. His research interests include image processing, computer vision, and pattern recognition.
\end{IEEEbiography}

\begin{IEEEbiography}[{\includegraphics[width=1in,height=1.25in,clip,keepaspectratio]{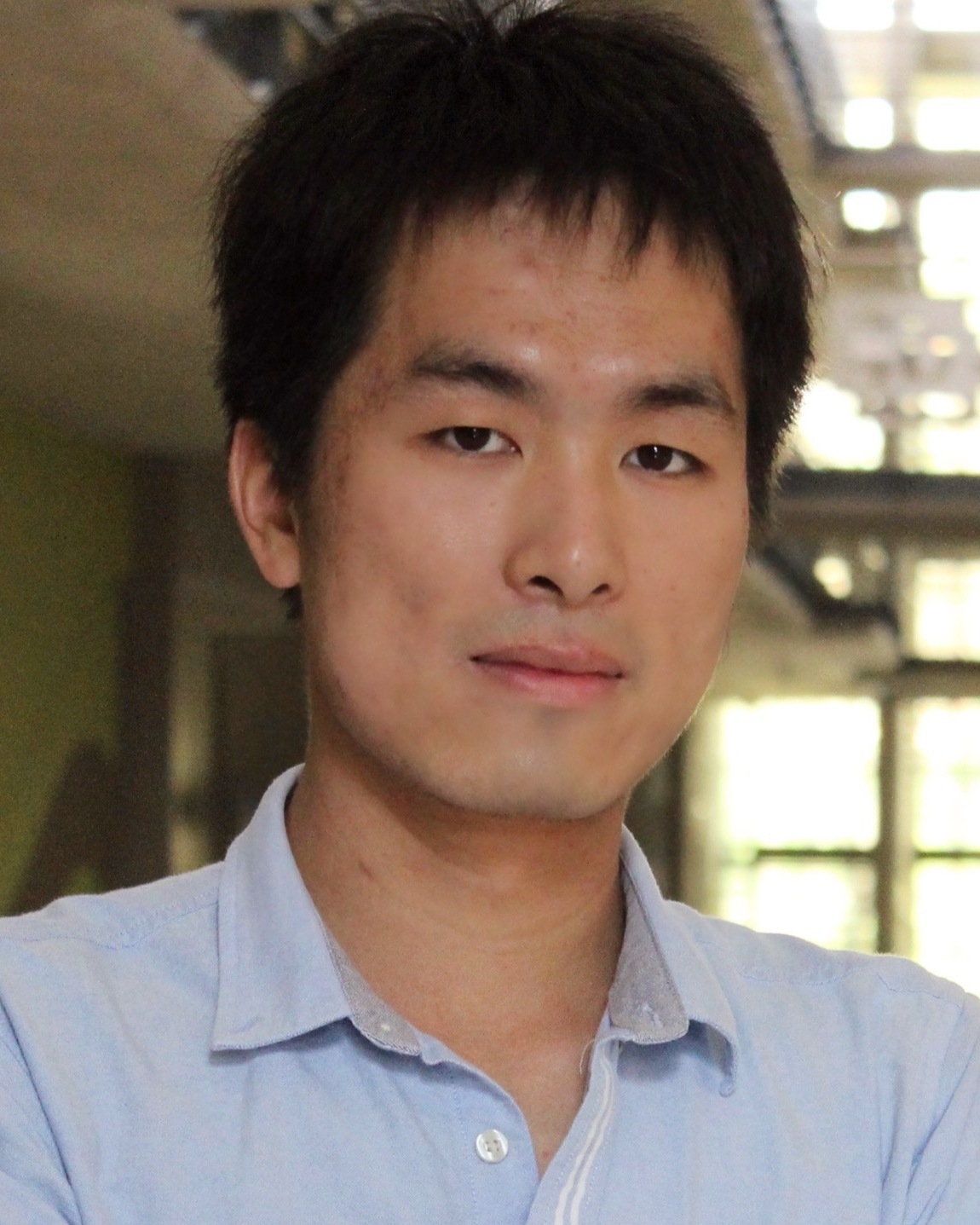}}]{Bing Zhu}(Member, IEEE) was born in Wuhan, P.R. China. He received his B.S. and Ph.D. degrees in Control Theory and Applications from Beihang University, Beijing, P.R. China, in 2007 and 2013, respectively. He was with University of Pretoria, Pretoria, South Africa, as a postdoctoral fellow supported by Vice-Chancellor Postdoctoral Fellowship from 2013 to 2015. He was with Nanyang Technological University, Singapore, as a research fellow from 2015 to 2016. He joined Beihang University, Beijing, P.R. China as an associate professor in 2016. 
Dr Zhu serves as an Associate Editor for Acta Automatica Sinica. His research interests include model predictive control, smart sensing for UAV and UGV, and demand-side management for new energy systems.

\end{IEEEbiography}

\end{document}